%% file: main.tex
\documentclass[10pt,twocolumn,letterpaper]{article}

\usepackage[pagenumbers]{cvpr} 
\usepackage{colortbl}
\definecolor{mygray}{gray}{.9}

\input{preamble}

\definecolor{cvprblue}{rgb}{0.21,0.49,0.74}

\def\paperID{****} 
\def\confName{CVPR}
\def\confYear{2024}
\newcommand{\name}{LaSagnA}

\title{\name: Language-based Segmentation Assistant for Complex Queries}

\author{Cong Wei$^{1,2*}$, Haoxian Tan$^{2*}$, Yujie Zhong$^{2}$$^{\dagger}$, Yujiu Yang$^{1}$$^{\dagger}$ and Lin Ma$^2$ \\
$^1$Tsinghua Shenzhen International Graduate School, Tsinghua University \\ $^2$Meituan Inc.\\
{\tt\small weic22@mails.tsinghua.edu.cn,} 
{\tt\small tanhx3@mail3.sysu.edu.cn,} \\
{\tt\small jaszhong@hotmail.com,} 
{\tt\small yang.yujiu@sz.tsinghua.edu.cn}
}

\begin{document}
\maketitle

\renewcommand{\thefootnote}{\fnsymbol{footnote}}
\footnotetext{$^*$Equal contribution.}
\footnotetext{$^{\dagger}$Corresponding authors.}

\input{sec/0_abstract}
\input{sec/1_intro}
\input{sec/2_related_work}
\input{sec/3_method}

\input{sec/4_exp}
\input{sec/5_conclusion}

{
    \small
    \bibliographystyle{ieeenat_fullname}
    \bibliography{main}
}


\end{document}

%% file: preamble.tex
\usepackage[dvipsnames]{xcolor}

\usepackage{graphicx}
\usepackage{amsmath}
\usepackage{amssymb}
\usepackage{booktabs}
\usepackage{bbding}
\usepackage{enumitem}
\usepackage[ruled]{algorithm2e}
\usepackage[dvipsnames]{xcolor}

\usepackage{multirow}
\usepackage{amsthm}
\usepackage{mathrsfs}
\usepackage{diagbox}
\usepackage{bm}
\usepackage{newfloat}
\usepackage{xspace}
\usepackage{xcolor}

\definecolor{citecolor}{RGB}{0, 113, 188}
\usepackage[pagebackref,breaklinks,colorlinks, citecolor=citecolor]{hyperref}

%% file: sec/0_abstract.tex
\begin{abstract}
Recent advancements have empowered Large Language Models for Vision (vLLMs) to generate detailed perceptual outcomes, including bounding boxes and masks. Nonetheless, there are two constraints that restrict the further application of these vLLMs: the incapability of handling multiple targets per query and the failure to identify the absence of query objects in the image. 
In this study, we acknowledge that the main cause of these problems is the insufficient complexity of training queries. Consequently, we define the general sequence format for complex queries. Then we incorporate a semantic segmentation task in the current pipeline to fulfill the requirements of training data. Furthermore, we present three novel strategies to effectively handle the challenges arising from the direct integration of the proposed format.
The effectiveness of our model in processing complex queries is validated by the comparable results with conventional methods on both close-set and open-set semantic segmentation datasets. Additionally, we outperform a series of vLLMs in reasoning and referring segmentation, showcasing our model's remarkable capabilities.
We release the code at \url{https://github.com/congvvc/LaSagnA}.

\end{abstract}

%% file: sec/1_intro.tex
\section{Introduction}\label{sec:intro}
Fueled by rapid progress in Large Language Models (LLMs), LLMs for vision (vLLMs) have emerged as a significant advancement\cite{liu2024visual,li2023blip,zhu2023minigpt,bai2023qwen}. By introducing a visual encoder and corresponding modality adapter, a pre-trained LLM can be converted into a powerful vLLM capable of generating textual responses based on input images. Recent studies\cite{Lai2023LISARS, Ren2023PixelLMPR, rasheed2023glamm, pi2023perceptiongpt} have made further progress in producing detailed perceptual outcomes, such as bounding boxes or masks, based on instructions and visual context. These developments are crucial for advanced applications, including intricate visual comprehension, interactive embodied agents, and localized content manipulation.


A line of research\cite{chen2023shikra,wang2024visionllm,peng2023kosmos} proposes to parse bounding boxes and segmentation masks as sequences of polygons. This enables the representation of masks as plain text, allowing end-to-end training within the existing LLM framework for vision. Most recent studies like LISA\cite{Lai2023LISARS, Ren2023PixelLMPR} have adopted a more simplified methodology. They utilize an additional special token to indicate the need for segmentation output. When the special token is present in the response, the corresponding features of the last layer of the vLLM are first projected to embeddings using MLP and then fed to an independent segmentation model (\eg SAM\cite{kirillov2023segment}) to decode the final masks. 


\begin{figure}[tb]
  \centering
   \includegraphics[width=\linewidth]{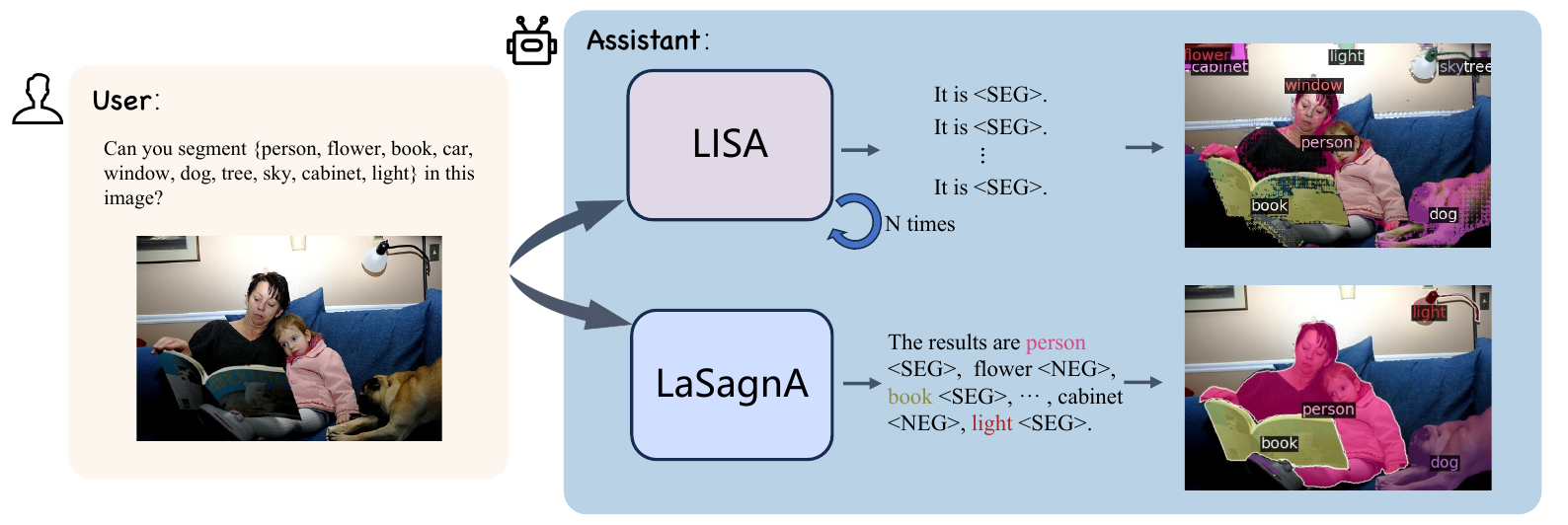}
  \vspace{-0.25in}
  \caption{The comparison between LISA\cite{Lai2023LISARS} and LaSagnA on complex query. LISA\cite{Lai2023LISARS} requires calling the model $N$ times to obtain the final result (where $N$ is the number of targets in the query), and it fails to identify non-existent categories such as \textit{window} and \textit{cabinet}. In contrast, LaSagnA can handle novel categories and accurately identify the existing targets in the image based on a single query.
  }
  \label{fig:compar}
\end{figure}

However, two drawbacks hinder the widespread use of this type of vLLMs: \textbf{1) the ineptitude to handle multiple targets in a single query}, and \textbf{2) the incapacity to identify whether the query objects/categories actually exist in the image}, as demonstrated in \cref{fig:compar}. 


\begin{figure*}[tb]
  \centering
   \includegraphics[width=\linewidth]{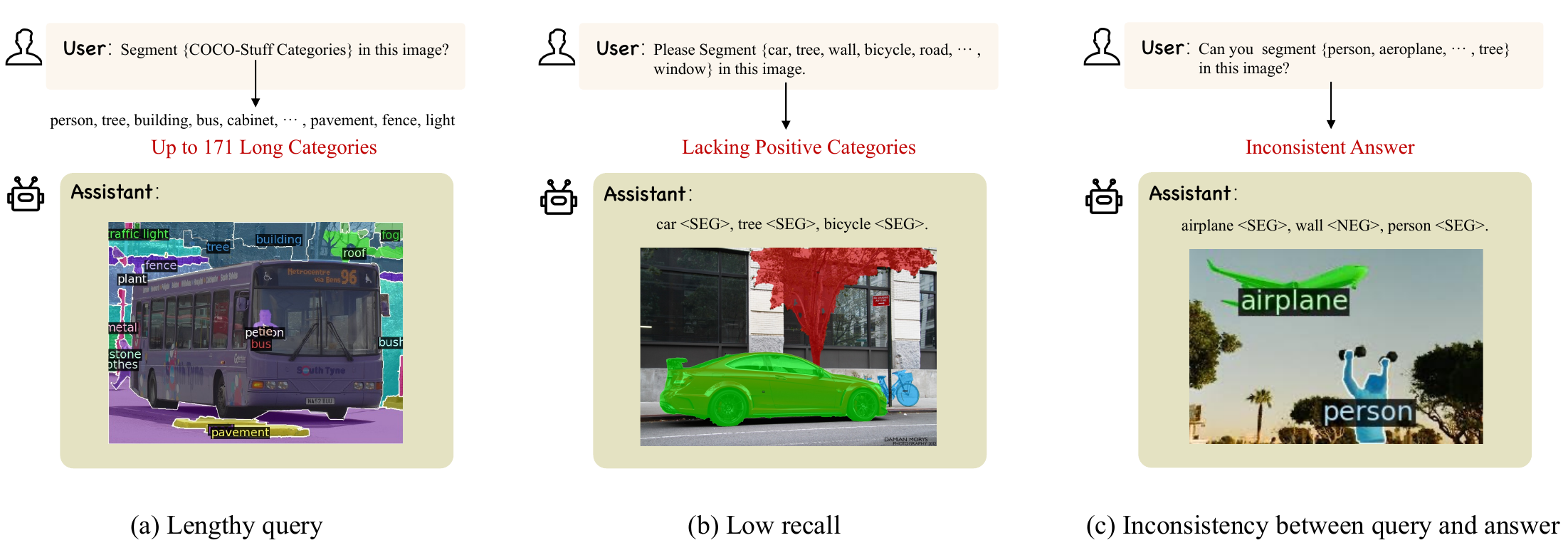}
  \vspace{-0.25in}
  \caption{The three problems existing in handling semantic segmentation tasks. (a) Lengthy input caused by a large number of categories in the dataset. (b) Low recall caused by incomplete sequence predictions. (c) Inconsistent category names between queries and responses.
  }
  \label{fig:3_problem}
\end{figure*}

\textbf{Firstly}, previous vLLMs (\eg LISA\cite{Lai2023LISARS}) are trained to recognize only one object per query. Therefore, it is compulsory to parse the query with multiple targets into individual queries and execute these queries separately. This process also involves complex post-processing to gather all the responses and generate the outcome. As the number of targets escalates, the efficiency of this entire process declines.
\textbf{Secondly}, these vLLMs assume the presence of query objects/categories in the image, which diverges from real-world scenarios. For instance, in scenarios that involve batch processing, the model is provided with the same list of categories for multiple images. If the model predicts masks for all the input categories regardless of their presence in the image, it may lead to a considerable number of false predictions. Additionally, the input query from users may not always be reliable and can sometimes include incorrect categories. Thus, the model should be capable of rejecting problematic requests to avoid generating erroneous signals for downstream applications.

In this study, our objective is to address the aforementioned issues. We acknowledge that the main cause of these problems lies in the insufficient complexity of training queries. 
Therefore, we introduce a general sequence format to incorporate multiple targets as well as non-existing targets into the queries during the training process. 
To train and evaluate the ability of vLLMs with respect to tackling complex queries, we discover that the semantic segmentation task naturally matches our requirements. Namely, it inherently requires the model to segment multiple categories simultaneously, without ensuring the presence of target categories in the image. 
Consequently, in this work, we opt to train and test the segmentation assistant using well-established semantic segmentation datasets.

Although the proposed sequence format allows the model to be trained on semantic segmentation datasets, effectively training the model to achieve high-quality results during inference remains a challenge. 
Further exploration reveals three main challenges: incomplete predictions, relatively long input sequences, and inconsistent category names between query and response, as shown in \cref{fig:3_problem}. To overcome these challenges, we present three corresponding strategies: sequence augmentation to handle incomplete predictions, random classes list to deal with lengthy inputs, and maintaining category order alignment with the query to resolve the issue of inconsistent responses.

The resultant vLLM is termed LLM-based Segmentation Assistant for Complex Queries (\name).
It is capable of supporting complex queries, namely, queries containing multiple arbitrary targets that may not even exist in the image. The proposed sequence format allows training the model using a large amount of manually labeled datasets, such as MS-COCO\cite{Lin2014MicrosoftCC} and ADE20K\cite{zhou2019semantic}. Different from LISA\cite{Lai2023LISARS}, this approach ensures more effective utilization of existing semantic segmentation data, thus equipping the model with a fundamental perception ability - performing semantic segmentation.


Extensive experiments demonstrate that \name~can nearly approach the performance of modern specialists (\eg Mask2Former\cite{cheng2022masked} and OVSeg\cite{liang2023open}) in both closed-set and open-set semantic segmentation. This inspiring result validates the capability of \name~in handling complex queries. Furthermore, \name~outperforms recently proposed vLLMs on referring segmentation and reasoning segmentation. 

In concurrent efforts, several studies\cite{zhan2023griffon,wu2023see,xia2023gsva} have been conducted to address the aforementioned challenges. Griffon\cite{zhan2023griffon} develops a novel dataset using GPT4V to encompass intricate scenarios. SESAME\cite{wu2023see} expands the existing refCOCO series by incorporating false-premise expressions generated by LLM, which feature comparable objects, attributes, or relations. Similar to our approach, GSVA\cite{xia2023gsva} incorporates multiple objects in the query and an extra token to indicate empty targets. However, it also includes supplementary data from gRefCOCO during training. In contrast, our methodology solely relies on the semantic segmentation dataset already blended in the LISA's pipeline.



In summary, our main contributions are three-fold.

- We identify a crucial limitation of recent vLLM-based segmentation assistants. These assistants struggle with handling queries that involve multiple arbitrary targets, which may or may not exist in the image. To overcome this limitation, we introduce a sequence format that takes into account multiple classes and negative classes. With the proposed format, the assistant can be readily trained on semantic segmentation datasets.

- To address the challenges associated with the training on the semantic segmentation task, we present three innovative techniques: random classes list, sequence augmentation, and order following. By employing these strategies, the vLLM model can effectively utilize segmentation datasets, thus significantly improving its overall segmentation performance.

- We conduct experiments on three distinct tasks and demonstrate the capability of the proposed model to handle complex queries. We reveal the potential of vLLM-based segmentation assistants in the fundamental perception task, namely, semantic segmentation. Moreover, we surpass a series of vLLMs in reasoning and referring segmentation.




%% file: sec/2_related_work.tex
\section{Related Work}
\subsection{Large Language Model for Vision}
The emergence of Large Language Model (LLM) architectures has significantly contributed to the development of vLMM. In this context, LLMs are enhanced with multimodal comprehension capabilities, allowing them to process both textual and visual information together\cite{li2023blip, alayrac2022flamingo, zhu2023minigpt, liu2024visual, liu2023improved, bai2023qwen}. A common approach in this framework is to integrate adapters, which facilitate the alignment of visual and textual representations within LLMs. Several notable examples of LLMs with multimodal comprehension include BLIP-2\cite{li2023blip}, Flamingo\cite{alayrac2022flamingo}, MiniGPT-4\cite{zhu2023minigpt}, LLaVA\cite{liu2024visual}, InstructBLIP\cite{liu2023improved}, and Qwen-VL\cite{bai2023qwen}. While these models have demonstrated improved performance in vision-language tasks, they solely produce textual outputs that describe the entire image. This restricts their applicability in tasks that require a more detailed understanding at the region or pixel level.

Several methods have been proposed to enhance Language Models with a more detailed comprehension capability\cite{chen2023shikra,wang2024visionllm,peng2023kosmos,you2023ferret,Lai2023LISARS,Ren2023PixelLMPR,rasheed2023glamm,pi2023perceptiongpt}.
Shikra\cite{chen2023shikra}, Ferret\cite{you2023ferret}, Kosmos-2\cite{peng2023kosmos}, and VisionLLM\cite{wang2024visionllm} are examples of models that provide grounding capabilities through regression of box coordinates of objects in their responses. Conversely, LISA\cite{Lai2023LISARS}, PixelLM\cite{Ren2023PixelLMPR}, GLaMM\cite{rasheed2023glamm}, and PerceptionGPT\cite{pi2023perceptiongpt} employ a mask decoder to predict object masks from special tokens. However, these models are trained on single-target datasets without the support of complex queries. Meanwhile, concurrent works\cite{zhan2023griffon,wu2023see,xia2023gsva} have been conducted to tackle complex queries with self-constructed data or data from other tasks. Nonetheless, our objective is to make up for this shortcoming of the model and enhance the overall performance using the existing semantic segmentation data. 


\subsection{Semantic Segmentation}
Semantic segmentation involves assigning class labels to each pixel in an image. Over the years, various designs\cite{long2015fully, noh2015learning, ronneberger2015u, chen2017deeplab, badrinarayanan2017segnet, yu2015multi, zhao2017pyramid, cheng2021per, lai2021semi, tian2022adaptive} have been proposed to effectively encode semantic information, including encoder-decoder architectures\cite{badrinarayanan2017segnet}, dilated convolutions\cite{yu2015multi}, pyramid pooling modules\cite{zhao2017pyramid}, and non-local operators\cite{zhu2019asymmetric}. These designs have significantly advanced traditional segmentation tasks. However, there is an increasing need to develop more generalized approaches for image segmentation. 

Recently, SAM\cite{kirillov2023segment} has demonstrated exceptional segmentation quality by utilizing billions of high-quality masks for training and showcased its versatility by supporting bounding boxes and points as input prompts. Other notable approaches like LSeg\cite{li2022languagedriven}, SimBaseline\cite{xu2022simple}, and OVSeg\cite{liang2023open} have explored the intersection of vision and language, enabling the prediction of masks based on on-the-fly language input. These studies primarily focus on employing language encoders to extract language features and then utilizing them in similarity computation between different modalities. In this work, we propose to empower the current vLLMs to deal with complex queries by incorporating a semantic segmentation task in the training pipeline. We also reveal that the extra task can benefit the existing high-level understanding tasks, \eg reasoning and referring segmentation.


%% file: sec/3_method.tex
\begin{figure*}[tb]
  \centering   \includegraphics[width=\linewidth]{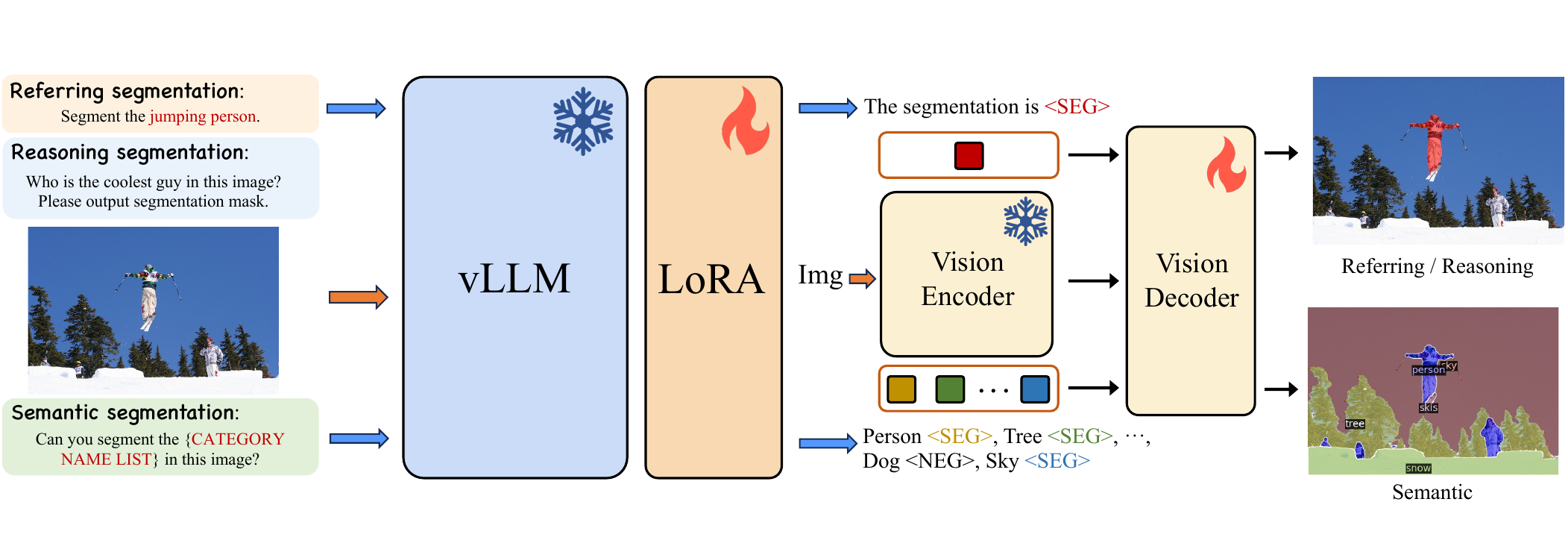}
   \vspace{-0.1in}
   \caption{Overview of \name. The vLLM generates a text response based on the instruction text and the input image. The vision encoder and decoder composite a standard SAM\cite{kirillov2023segment} which is trained to predict a mask based on the textual embedding generated by the vLLM. We only finetune the vLLM using LoRA and train the decoder of SAM.}
   \label{fig:model}
\end{figure*}

\section{Method}

 In this section, we offer a summary of vLLM-based segmentation assistants in \cref{subsec:recap}. We then examine limitations in previous methods in \cref{subsec:problem}. Next, we focus on discussing the format of our input sequence in \cref{subsec:sequence}. Lastly, a comprehensive explanation of our proposed training strategies in the training recipe is detailed in \cref{subsec:training}.

\subsection{vLLM-based Segmentation Assistants}\label{subsec:recap}


\noindent\textbf{Overall architecture.}\quad Since the design of network architecture is not the focus of this work, we build \name~following LISA\cite{Lai2023LISARS}, as depicted in \cref{fig:model}. The vLLM is responsible for generating a text response based on the instruction text and the input image. The vision encoder and decoder follow a standard SAM\cite{kirillov2023segment} architecture, which takes an image as input and generates a mask based on the input prompt. In our case, the prompt refers to the textual feature from the vLLM.

\noindent\textbf{Training objective.}\quad The model can be trained jointly on multiple tasks using unified losses as all tasks are modeled as language instruction. We employ an autoregressive cross-entropy loss for next token prediction, a per-pixel binary cross-entropy loss, and a DICE loss for mask supervision. The overall objective constitutes the weighted sum of these losses, calibrated by $\lambda_{bce}$ and $\lambda_{dice}$:
\begin{equation}
\label{eq:total_loss}
\mathcal{L}= \mathcal{L}_{text}+\lambda_{bce}\mathcal{L}_{bce}+\lambda_{dice}\mathcal{L}_{dice}.
\end{equation}

\noindent\textbf{Advantages of vLLM-based segmentation assistants.}
In comparison with the conventional segmentation methods such as UpperNet~\cite{xiao2018unified} and Mask2Former~\cite{cheng2022masked} which only adopt classifiers for categorization, vLLMs can understand abstract natural language instructions. Therefore, vLLMs are good at referring or reasoning segmentation tasks in general, as they are capable to parse the meaning of the query sentence and locate the target objects/categories accordingly.

\subsection{Limitations of vLLM-based Assistants}\label{subsec:problem}
We argue that the sub-optimal performance of current vLLMs can be attributed to the strong prior implied by their training process, where \textbf{all training queries solely comprise a single target that already exists in the image}. To be specific, the training segmentation dataset consists of reference segmentation datasets (refCOCO series\cite{yu2016modeling}), reasoning segmentation datasets\cite{Lai2023LISARS}, and semantic segmentation datasets (ADE20K\cite{zhou2019semantic} and COCO-stuff\cite{caesar2018coco}). The first two datasets are specifically designed for querying a single target in the image, while the semantic datasets are also utilized similarly. An existent category is randomly selected to construct the training query, and the corresponding binary segmentation mask is employed as the ground truth for providing mask loss supervision. Given these settings, it is unreasonable to expect the model to effectively handle complex queries that involve multiple targets and nonexistent categories.

\subsection{Learning with Complex Sequence}
\label{subsec:sequence}
\noindent\textbf{The proposed sequence format.}
The complex query is characterized by multiple targets and the absent categories in the image. To satisfy this requirement, it is intuitive to include more categories in the training query and prompt the model to return existent categories in the response. Consequently, we can define the following template:

\vspace{0.05in}\texttt{\textbf{USER}: <IMAGE> Can you segment the \{CATEGORY NAME LIST\} in this image?} 
\vspace{0.05in}
\texttt{\textbf{ASSITANT}: \{P-CAT\}<SEG>,...,\{P-CAT\}<SEG>}.\vspace{0.1in}

\noindent Here, \texttt{<IMAGE>} denotes the placeholder for tokens of image patches. \texttt{\{CATEGORY NAME LIST\}} represents the classes list of dataset. \texttt{<SEG>} is a newly introduced token used for decoding the masks of positive classes, following the LISA design. \texttt{\{P-CAT\}} denote the names of the existent classes in the image. The number of the combination of "\texttt{\{P-CAT\}<SEG>}" matches that of the existent categories in the training image.

\noindent\textbf{Incoorperating semantic segmentation task.}\quad
Then the associated training data is required to support complex queries. Given a predefined categories list, the typical semantic segmentation task involves identifying the classes in the image and their corresponding pixels, which naturally aligns with our need for multiple targets and nonexistent classes. Therefore, in order to enhance vLLM's capability in handling complex queries, we choose to incorporate semantic segmentation tasks into the training paradigm as well as the evaluation.

\noindent\textbf{Problems of training on semantic segmentation datasets.}\quad
Although the current training pipeline for \name~can seamlessly integrate semantic segmentation tasks with the proposed sequence construction framework, directly training the model with this format yields sub-optimal performance. This unexpected outcome arises from three major issues: incomplete predictions, lengthy input sequences, and inconsistent category names between query and response, as is shown in \cref{fig:3_problem}. We discuss them in detail as follows.

1) \textbf{Incomplete predictions.}\quad In practice, it has been observed that the model often fails to predict all classes, leading to a significant impact on its recall. This deficiency can be attributed to uncertainties in recognizing certain classes in the image. Additionally, the incorporation of training sequences with different lengths further complicates the model optimization.

2) \textbf{Lengthy input sequences.}\quad The length of the input sequence is highly influenced by the number of classes, as we include all the categories from the sampled dataset in the query. A longer list of class names consumes more tokens, thereby limiting the capacity for generating new tokens and impacting the recall of semantic segmentation prediction. Moreover, longer training sequences necessitate additional GPU memory and extended training time. In reality, many categories are absent in the image, making it unnecessary to include an excessive number of categories.

3) \textbf{Inconsistent category names.}\quad The model aims to generate predictions based on the target list specified in the query. However, in the context of open-set segmentation scenarios, we have observed that the model often produces category names that are found in the training data instead of the ones specified.
 




\subsection{Training Recipe on Complex Queries}\label{subsec:training}
To mitigate the above-mentioned problems, we propose a training recipe that significantly enhances the training quality on semantic segmentation datasets thus improving the segmentation performance.
The recipe contains three strategies, which are elaborated in the following.


\noindent\textbf{Sequence augment.}\quad 
To enhance recall affected by incomplete predictions, 
one possible method to address incomplete predictions is to artificially reduce the likelihood of sampling the \texttt{EOS} token and extend the length of response. Nonetheless, this approach often leads to noisy and duplicated predictions, with potential side effects on other tasks. To overcome these challenges, we enhance the response in the training sequence by incorporating the negative classes from the provided list of classes. The predicted class names in the response match the class names in the query. The model predicts all the classes mentioned in the query, with positive predictions denoted by the special token \texttt{<SEG>} and negative predictions denoted by \texttt{<NEG>}. This modification offers three main advantages. First, it encourages the generation of longer sequences with a greater number of classes. Second, it explicitly instructs the model to recognize the negative classes. Last but not least, it standardizes the response to a relatively fixed length,  facilitating model learning.

\noindent\textbf{Random classes list.}\quad 
To address the problem of lengthy sequences, we suggest randomly sampling the class list in the query. 
Specifically, we randomly select a number of targets from the complete list of category names, without considering if they encompass all existent objects in the image. By maintaining a dynamic list during training, our model becomes adept at handling various targets specified in the inference query, thereby enabling open-world segmentation. Moreover, this approach facilitates chunk inference, where lengthy target lists can be partitioned into smaller queries for parallel execution.

\noindent\textbf{Target order consistency.}\quad 
To enhance class consistency between the query and response during inference, we ensure that the order of the target categories in the response aligns with that in the question. The presence of similarity between the question and answer in the training dataset encourages the model to generate a response that relies more on the query context.

In line with the aforementioned strategies, the final input format for the semantic segmentation task is as follows:

\vspace{0.1in}\texttt{\textbf{USER}: <IMAGE> Can you segment the \{SAMPLED NAME LIST\}in this image?}

\texttt{\textbf{ASSITANT}: \{P-CAT\}<SEG>,\{N-CAT\}<NEG>,\\...,\{N-CAT\}<NEG>,\{P-CAT\}<SEG>.}\vspace{0.1in}

\noindent Here, \texttt{\{SAMPLED NAME LIST\}} represents a list of sample classes, which consists of random classes from the training data. The newly introduced token, \texttt{<SEG>}, is used for decoding the masks of positive classes, while \texttt{<NEG>} serves as a token to indicate negative classes.  \texttt{\{P-CAT\}} and \texttt{\{N-CAT\}} are used to represent the names of the existent classes and nonexistent classes, respectively. The total number of classes in the response corresponds to the classes listed in the query. For our implementation, we have constructed various templates using the above format. The templates we used will be further detailed in the appendix.

%% file: sec/4_exp.tex
\section{Experiments}

\begin{table*}[t]
  \centering
  \begin{tabular}{c|l|l|c|c|c}
    \toprule[1.1pt]
    Type & Method & Backbone & ADE20K & COCO-Stuff & Cityscapes\\
    \midrule[0.7pt]
    \multirow{6}{*}{\shortstack{Segmentation Specialist}}  & 
    RefineNet\cite{lin2019refinenet}& R101 & 40.2 & 33.6 & 73.6 \\
    & Upnet\cite{xiao2018unified}& R50 & 41.2 & - & - \\
    & OCR\cite{YuanCW19}& HRNetV2 & 45.6 & 40.5 & 81.6 \\
    & Mask2former\cite{cheng2022masked}& R50 & 47.2 & - & 79.4 \\
    & SegViT\cite{zhang2022segvit}& ViT-L & 54.6 & 50.3 & - \\
    & MaskDINO+\cite{li2023mask}&SwinL & 56.6& - & -\\
    \midrule[0.5pt]
    \multirow{2}{*}{\shortstack{vLLM-based Assistant}}
    & LISA-7B\cite{Lai2023LISARS} & ViT-H& 19.7 & 24.4 & 38.5 \\
    & \textbf{\name~(ours)} & ViT-H & 42.0 & 43.9 & 63.2\\
    \bottomrule[1.1pt]
  \end{tabular}
    \caption{ Quantitative results (mIoU at single scale) on closed-set semantic segmentation. 
  }
  \label{tab:semseg}
\end{table*}

\begin{table*}[t]
  \centering
    \begin{tabular}{c|l|c|c|c|c}
    \toprule[1.1pt]
    Type & Method  & PC-459 & A-150 & PC-59 & PAS-20 \\
    \midrule[0.7pt]
    \multirow{9}{*}{\shortstack{Segmentation Specialist}} 
    & LSeg\cite{li2022language} & - & - & - & 47.4  \\
    & LSeg+\cite{ghiasi2022scaling} & 5.2 & 13.0 & 36.0 & - \\
    & GroupViT\cite{xu2022groupvit} & 4.9 & 10.6 & 25.9 & 50.7 \\
    & PACL\cite{mukhoti2023open} & - & 31.4 & 50.1 & 72.3 \\
    & ZegFormer\cite{ding2022decoupling} & 10.4  & 18.0 &45.5 & 89.5 \\
    & SimBaseline\cite{xu2022simple} & - & 20.5 &47.7 & 88.4\\
    & DeOP\cite{Han2023ZeroShotSS} & 9.4 & 22.9 & 48.8 & 91.7\\
    & OVSeg\cite{liang2023open} & 11.0  &24.8 &53.3 & 92.6\\
    & SAN\cite{xu2023side} & 12.6  &27.5 &53.8 & 94.0\\
    \midrule[0.5pt]
    \multirow{2}{*}{\shortstack{vLLM-based Assistant}} 
    & \textbf{\name$^\dagger$~(ours)} & 8.5 & 14.3 & 46.1 & 69.8\\
    & \textbf{\name~(ours)} & 9.8 & \cellcolor{mygray}42.0 & 39.6 & 61.8 \\
    \bottomrule[1.1pt]
  \end{tabular}
    \caption{ Quantitative results on open-set semantic segmentation. 
    \name~reported here utilizes ADE20K data and hereby achieves a high number on A-150.
    $^\dagger$ denotes the model is trained on COCO-Stuff only for fair comparison.
    }
  \label{tab:openseg}
\end{table*}

\begin{table*}[tb]
  \centering
    \begin{tabular}{c|l|ccc|ccc|cc|c}
    \toprule[1.1pt] 
    \multirow{2}{*}{ Type } &\multirow{2}{*}{ Method } & \multicolumn{3}{c|}{ refCOCO } & \multicolumn{3}{c|}{ refCOCO+ } & \multicolumn{2}{c|}{ refCOCOg } & reasonSeg\\
    \cline { 3 - 11 } & & val & testA & testB & val & testA & testB & val(U) & test(U) & val\\
    \midrule[0.7pt]
    \multirow{6}{*}{\shortstack{Segmentation\\Specialist}} & MCN\cite{luo2020multi}  & 62.4 & 64.2 & 59.7 & 50.6 & 55.0 & 44.7 & 49.2 & 49.4  & -\\
    & VLT \cite{ding2021vision}& 67.5 & 70.5 & 65.2 & 56.3 & 61.0 & 50.1 & 55.0 & 57.7  & -\\
    & CRIS\cite{wang2022cris} & 70.5 & 73.2 & 66.1 & 62.3 & 68.1 & 53.7 & 59.9 & 60.4  & -\\
    & LAVT\cite{yang2022lavt} & 72.7 & 75.8 & 68.8 & 62.1 & 68.4 & 55.1 & 61.2 & 62.1 & -\\
    & ReLA\cite{liu2023gres}  & 73.8 & 76.5 & 70.2 & 66.0 & 71.0 & 57.7 & 65.0 & 66.0 & -\\
    & PolyFormer-B\cite{liu2023polyformer} & 74.8 & 76.6 & 71.1 & 67.6 & 72.9 & 59.3 &67.8 &69.1 & -\\
    \midrule[0.5pt] 
    \multirow{6}{*}{\shortstack{vLLM-based\\Assistant}} & LISA-7B\cite{Lai2023LISARS} & 74.1 & 76.5 & 71.1 & 62.4 & 67.4 & 56.5 &  66.4 & 68.5 & 46.0\\
    & Pixellm-7B\cite{Ren2023PixelLMPR}& 73.0 &76.5 & 68.2 & 66.3 & 71.7 & 58.3 &69.3 & 70.5 & -\\
    & PerceptionGPT-7B\cite{pi2023perceptiongpt}& 75.1 &78.6 &71.7 &68.5 &73.9 &61.3 &70.3 &71.7 & -\\
    & SESAME\cite{wu2023see}& 74.7 & - & - & 64.9 & - & - & 66.1 & - & - \\
    & GSVA-7B\cite{xia2023gsva}& 76.4 &77.4 &72.8 &64.5 &67.7 & 58.6 & 71.1 & 72.0 & -\\
    & \textbf{\name~(ours)} & 76.8 & 78.7 & 73.8 & 66.4 & 70.6 & 60.1 & 70.6 & 71.9 & 47.2\\
    \bottomrule[1.1pt]
    \end{tabular}
  \caption{ Quantitative results (cIoU) on multiple high-level understanding tasks including three conventional referring segmentation benchmarks (refCOCO series) and reasoning segmentation dataset proposed by LISA\cite{Lai2023LISARS}. 
  }
\label{tab:high-level}
\end{table*}

\begin{table*}[t]
  \centering
    \begin{tabular}{c|l|c|cc|cc|cc}
    \toprule[1.1pt] 
    \multirow{2}{*}{ Type } & \multirow{2}{*}{ Method } & \multirow{2}{*}{ Zero-shot } & \multicolumn{2}{c|}{ val } & \multicolumn{2}{c|}{ testA } & \multicolumn{2}{c}{ testB } \\
    \cline { 4 - 9 } & & & cIoU & gIoU & cIoU & gIoU & cIoU & gIoU \\
    \midrule[0.7pt]
    \multirow{4}{*}{\shortstack{Segmentation Specialist}} & MattNet\cite{yu2018mattnet}  & \XSolidBrush & 47.5 & 48.2 & 58.7 & 59.3 & 45.3 & 46.1 \\
    & LTS \cite{jing2021locate}& \XSolidBrush & 52.3 & 52.7 & 61.9 & 62.6 & 49.9 & 50.4 \\
    & CRIS\cite{wang2022cris} & \XSolidBrush & 55.3 & 56.3 & 63.8 & 63.4 & 51.0 & 51.8 \\
    & LAVT\cite{yang2022lavt} & \XSolidBrush & 57.6 & 58.4 & 65.3 & 65.9 & 55.0 & 55.8 \\
    
    \midrule[0.5pt] 
    \multirow{3}{*}{\shortstack{vLLM-based Assistant}} & LISA-7B\cite{Lai2023LISARS} & \XSolidBrush & 38.7 & 32.2 & 52.6 & 48.5 & 44.8 &  39.7 \\

    & GSVA-7B\cite{xia2023gsva}& \XSolidBrush & 61.7 &63.3 &69.2 &70.1 & 60.3 & 61.3 \\


    & \textbf{\name~(ours)} &  \Checkmark & 38.1 & 32.4 & 50.4 & 47.3 & 42.1 & 38.9 \\
    \bottomrule[1.1pt]
    \end{tabular}
  \caption{ Quantitative results (cIoU and gIoU) on gRefCOCO. Our method also achieves promising zero-shot performance, which is comparable to LISA-7B\cite{Lai2023LISARS} that incorporates gRefCOCO during training.
  }
\label{tab:gref}
\end{table*}

\subsection{Implementation Details}
\noindent\textbf{Architecture.}\quad \name~comprises a vLLM and a general segmentation model. To enable efficient training, we utilize a pre-trained LLaVA-7B\cite{liu2024visual} as the large language model for vision. As for an independent segmentation model, we adopt SAM\cite{kirillov2023segment} with ViT-H backbone. For the training stage, LoRA\cite{hu2021lora} is employed to finetune the LLM efficiently. Additionally, we train the mask decoder in SAM, while keeping all other parameters frozen to preserve their original capabilities.

\noindent\textbf{Training data.}\quad The entire training dataset consists of four different types of tasks. ADE20K\cite{zhou2019semantic}, COCO-Stuff\cite{caesar2018coco}, and OpenImage\cite{openimages} are chosen for semantic segmentation without specific explanation. We select refCLEF\cite{kazemzadeh2014referitgame} and the refCOCO series\cite{yu2016modeling} for referring segmentation. We use the same reasoning segmentation data as LISA\cite{Lai2023LISARS}. Following the practice of previous vLLMs, we also include LLAVA-150k\cite{liu2024visual} for the VQA task. To train jointly on these tasks, we sample data from a randomly selected dataset for each data loader process during each iteration, which is expected to optimize multiple tasks concurrently within a single iteration.

\noindent\textbf{Hyperparameters.}\quad The training process lasts approximately 50 hours using 8 NVIDIA A100 GPUs. We employ the AdamW optimizer with a learning rate of 3e-4 and incorporate the WarmupDecayLR learning rate scheduler with a warmup steps count of 100. The batch size is set to 2 on each device. The model is trained for approximately 1 epoch (18000 steps). Additionally, the modulating parameters $\lambda_{bce}$ and $\lambda_{dice}$ in \cref{eq:total_loss} are assigned values of 1.0 and 0.5, respectively. 

\noindent\textbf{Evaluation metrics.}
The widely used metric, mean Intersection-over-Union (mIoU), is employed for evaluating semantic segmentation tasks. Consistent with previous studies in referring segmentation, we adopt another two metrics: cumulative Intersection-over-Union (cIoU) and the average of all
per-image Intersection-over-Unions (gIoU). It should be noted that cIoU tends to favor large-area objects due to its calculation method, which involves the cumulative intersection over the cumulative union.

\subsection{Comparisons with State-of-the-Arts}


We evaluate the effectiveness of the proposed model by comparing it with other conventional methods on closed-set and open-set semantic segmentation datasets, as is shown in \cref{tab:semseg} and \cref{tab:openseg}. Furthermore, we present the results of referring segmentation in both traditional and generalized settings in \cref{tab:high-level} and \cref{tab:gref} respectively, to highlight the improvement achieved in high-level vision-language tasks.

\noindent\textbf{Closed-set semantic segmentation results.}
The performance of \name~on closed-set semantic segmentation tasks is illustrated in \cref{tab:semseg}. It significantly outperforms the vLLM-based baseline with improvements of +22.3 on ADE20K, +19.5 on COCO-Stuff, and +24.7 on Cityscape. Furthermore, our model achieves performance that is close to that of RefineNet\cite{lin2019refinenet}, further validating its capability to handle complex queries.

The vLLM-based architecture differs from conventional segmentation specialists by emphasizing high-level understanding rather than capturing low-level visual features. Additionally, our framework utilizes an independent segmentation model that is primarily designed for geometric prompts such as boxes and points, as opposed to descriptive text. As a result of these design choices, there are limitations in accurately identifying small or crowded targets and generating precise masks when compared to specialized segmentation experts.

In addition, multi-task training, if not carefully tuned, can harm individual performance. When we follow the multi-dataset training approach of \name~to train Mask2-former\cite{cheng2022masked}, we observe an obvious decrease in performance (from 47.2 to 34.9 on ADE20K, from 79.4 to 66.1 on Cityscapes).

Considering all the aforementioned factors, our model still lags behind the current state-of-the-art (SOTA) segmentation specialists\cite{li2023mask,zhang2022segvit}, although it represents a significant advancement in the field of vLLM-based assistants.



\begin{table*}[tb]

\centering
\begin{tabular}{l|c|c|c|ccc|c}
\toprule[1.1pt]
\multirow{2}{*}{Components} & ADE20K & COCO-Stuff & Cityscapes & \multicolumn{3}{|c|}{ refCOCO }  & reasonSeg\\
\cline { 2 - 8 } & val & val & val & val & testA & testB  & val \\

\midrule[0.2pt]
LISA & 19.7 & 24.4 & 38.5 & 74.1 & 76.5 & 71.1 & 46.0 \\
+ class list &  35.2 & 40.5 & 59.9 & 73.9 & 76.6& 70.4 & 46.3  \\
+ random question  & 37.4 & 42.0 & 59.9 & 73.9 & 76.2 & 70.6 & 49.5 \\
+ sequence aug & 37.6 & 42.4& 60.1  & 73.5 & 76.8 & 71.1 & 52.7\\
+ target order consistency (\name) & 38.6 & 43.2 & 61.3 & 74.5 & 77.0 & 71.2 & 54.0 \\
\bottomrule[1.1pt]
\end{tabular}
\caption{Ablation study of each proposed component. We show the results of integrating different strategies into the baseline. All ablation studies follow the same training setting as LISA\cite{Lai2023LISARS}.}
\label{tab:components}
\end{table*}

\begin{figure*}[tb]
  \centering
  \includegraphics[width=\linewidth]{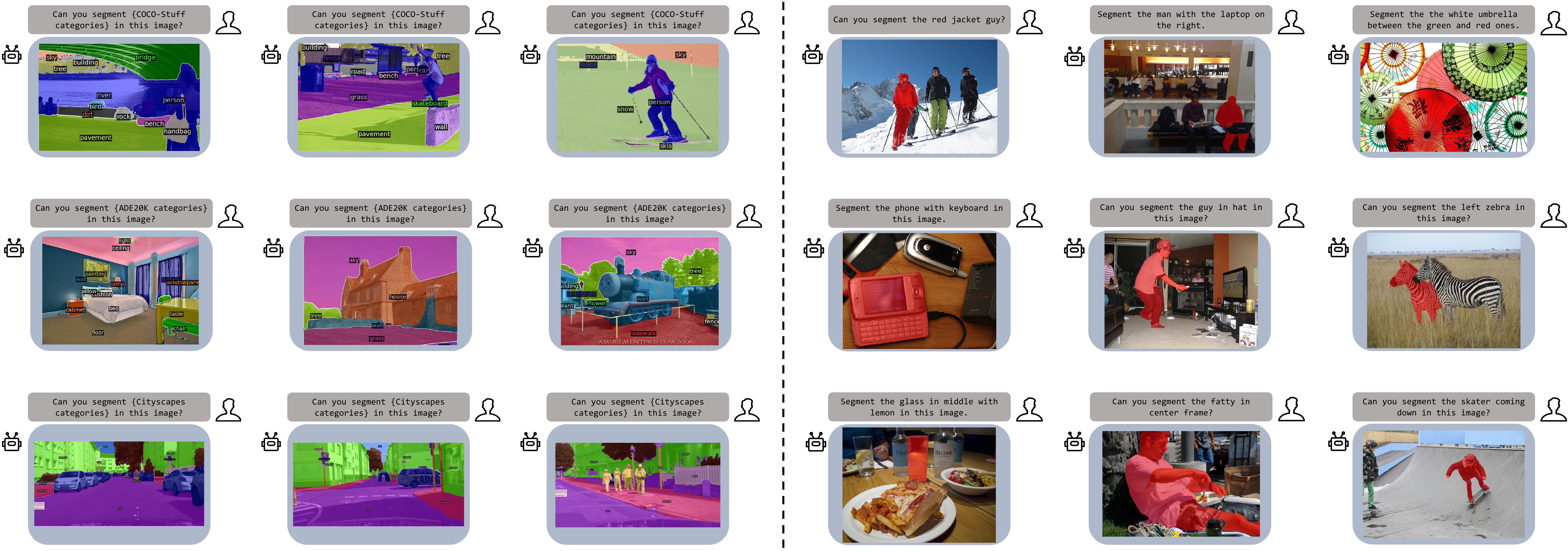}
  \caption{Qualitative results of \name’s performance on complex queries and single object scenarios.}
  \label{fig:vis_sem}
\end{figure*}

\noindent\textbf{Open-set semantic segmentation results.}
\name~can be seamlessly extended to open-set scenarios without any network architecture modifications. We simply task \name~with predicting the categories provided in the question. The results obtained by our models, as presented in \cref{tab:openseg}, are comparable to those achieved by other models that employ complex cross-modal alignment techniques and custom network designs. Specifically, our full model achieves mIoU scores of 9.8 on PC-459, 39.6 on PC-59, and 61.8 on PAS-20. However, since the full model has been trained on ADE20K data, its performance on A-150 does not accurately reflect its open-set capabilities. Therefore, for a fair comparison, we also train the model solely on COCO-Stuff, which yields mIoU scores of 8.5 on PC-459, 14.3 on A-150, 46.1 on PC-59, and 69.8 on PAS-20 respectively. 

\noindent\textbf{Referring and reasoning segmentation results.}
We demonstrate the efficacy of our model on both the traditional single-target referring segmentation and the generalized referring segmentation benchmark.

Despite the refCOCO series' focus on single target segmentation, \name~achieves commendable performance. As presented in Table~\ref{tab:high-level}, our model attains the highest cIoU scores across most metrics, especially excelling in the more challenging benchmarks such as reasonSeg (47.2) and refCOCOg val(U) (70.6). By incorporating negative classes into semantic segmentation tasks, the model is directed toward identifying additional concepts, thereby enhancing the performance of traditional referring tasks.

The generalized referring expression segmentation benchmark, gRefCOCO, encompasses both multi-target and non-target expressions, aligning well with our design. Our model’s performance, as demonstrated in \cref{tab:gref}, exhibits promising results even in \textbf{zero-shot} scenarios, comparable to the LISA-7B\cite{Lai2023LISARS} that incorporates gRefCOCO during training.

\begin{figure}[h]
  \centering
  \includegraphics[width=\linewidth]{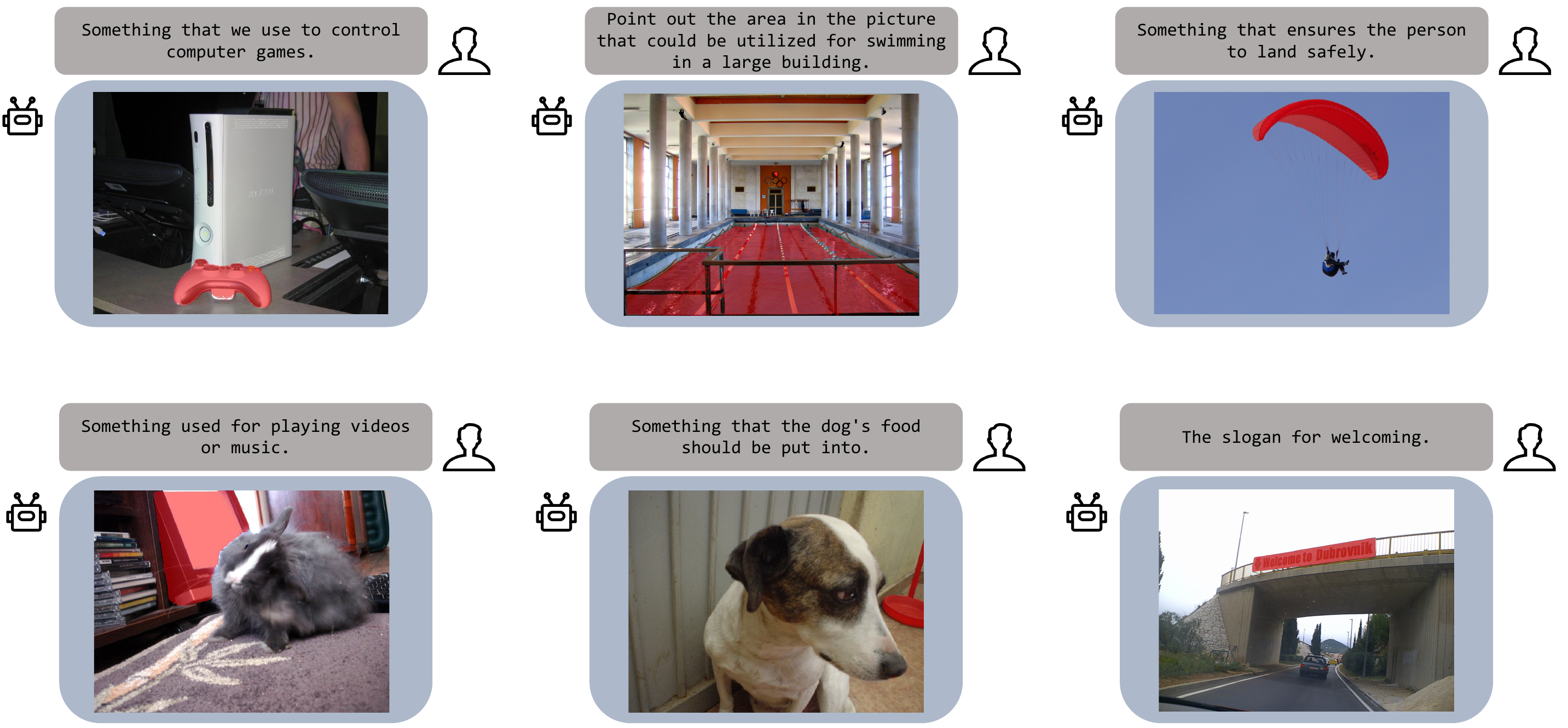}

  \caption{Qualitative results of \name’s performance on reasoning segmentation.}
  \label{fig:vis_reason}
\end{figure}

\noindent\textbf{Visualization.}\quad We provide visualization for the supported scenarios of \name, including arbitrary multi-object query and single-target referring query in \cref{fig:vis_sem} and reasoning query in
\cref{fig:vis_reason}. The visualization results demonstrate that the proposed \name~could utilize the powerful understanding capability of vLLM along with our specifically designed strategies to conduct multiple high-level understanding tasks simultaneously.


\subsection{Ablation Studies}
To perform a thorough ablation study, we assess the different variations using various benchmarks, including semantic segmentation, referring segmentation, and reasoning segmentation. To expedite the experiments, we omit OpenImage data from the semantic segmentation task and maintain a consistent number of training iterations as in LISA\cite{Lai2023LISARS}.




\noindent\textbf{Effectiveness of the proposed components.} To evaluate the effectiveness of our proposed strategies, we have conducted a series of experiments on each component in \cref{tab:components}. Expanding to a list of query classes significantly enhances the performance (+15.5 on ADE20K, +16.1 on COCO-Stuff, and +21.4 on Cityscapes). Randomly sampling the subset of all categories in the user query and augmenting the answer with negative classes could further improve the performance on both semantic segmentation and reasoning segmentation tasks. By ensuring the same order of categories in the answer as in the question, we ultimately achieve a performance of 38.6 on ADE20K, 43.2 on COCO-Stuff, and 61.3 on Cityscapes.

It is worth noting that the performance of referring segmentation exhibits only minor variations upon incorporating each proposed strategy. This demonstrates the stability of our model in handling multiple segmentation tasks simultaneously. Another possible explanation is that the proposed methods primarily focus on improving the performance of semantic and multi-target segmentation, rather than referring/reasoning segmentation.


\noindent\textbf{Inference chunk size of categories.}\quad 
\name~maintains a dynamic random list of category names during the training stage for semantic segmentation tasks. Consistent with the training strategies,  we partition the class names of each dataset into multiple chunks during inference. In \cref{tab:chunksize}, we examine the performance when different chunk sizes are used in the inference stage. Using a small chunk size could severely deteriorate the performance of semantic segmentation both in ADE20K and COCO-Stuff benchmarks.  Small chunk size needs more turns of conversation to accomplish semantic segmentation involving multiple categories for a single image, which may increase the uncertainty of vLLM's answer. By increasing the chunk size from $N/4$ to $N$, we achieve enhanced performance with mIoU improvements of 11.0 and 9.6 on ADE20K and COCO-Stuff, respectively.

\begin{table}[t]

  \centering
    \begin{tabular}{c|c|c}
    \toprule[1.1pt]
    Chunk Size   & ADE20K & COCO-Stuff \\
    \midrule[0.7pt] 
    $N/4$ & 31.0 & 34.3  \\
    $N/3$  & 30.8 & 33.2  \\
    $N/2$   & 38.9 & 41.8 \\
    $N$ & 42.0 & 43.9   \\
    \bottomrule[1.1pt]
  \end{tabular}
    \caption{ The influence of inference chunk size of categories. $N$ denotes the size of the whole categories on the current dataset.}
\label{tab:chunksize}
\end{table}


%% file: sec/5_conclusion.tex
\section{Conclusion and Future work}

In this study, we attempt to overcome the limitations of Large Language Models for vision (vLLMs) in handling complex queries. We introduce an input format specifically designed for complex queries and propose the integration of the semantic segmentation task into the training process to enhance support for such queries. To effectively tackle the challenges associated with the integration of this format, we propose three novel strategies. Our model, known as Language-based Segmentation Assistant for Complex Queries (\name), demonstrates its remarkable abilities on three distinct tasks.

In the future, we plan to expand and improve the capabilities of vLLMs. Specifically, we will explore the integration of more complex perception tasks and datasets such as panoptic segmentation. Additionally, we will work towards developing a lighter and more efficient vLLM and mask decoder.